\documentclass[letterpaper, 10 pt, conference]{ieeeconf}  

\IEEEoverridecommandlockouts                             
\usepackage{graphics}
\usepackage{epsfig} 
\usepackage{mathptmx} 
\usepackage{times} 
\usepackage{amsmath} 
\usepackage{amssymb}  
\usepackage{graphicx}
\usepackage{float}
\usepackage{tabularray}
\usepackage{bm}
\usepackage[dvipsnames]{xcolor}
\usepackage[caption=false]{subfig}
\usepackage{mathtools}
\usepackage{stfloats}
\usepackage{mathrsfs}
\usepackage{multirow}
\usepackage{makecell}
\usepackage{vcell}
\usepackage{booktabs}
\usepackage{diagbox}
\usepackage{csquotes}
\usepackage{cite}
\usepackage{tikz}
\usepackage{lipsum}
\usepackage{schemata}

\usepackage[hyphens]{url}
\usepackage[hidelinks]{hyperref}
\hypersetup{breaklinks=true}
\usepackage{cite}

\usepackage{algorithm}
\usepackage[noend]{algpseudocode}

\algrenewcommand\algorithmicforall{\textbf{for each}}
\algrenewcommand\algorithmicindent{.8em}

\algdef{SE}[DOWHILE]{Do}{doWhile}{\algorithmicdo}[1]{\algorithmicwhile\ #1}%
\algnewcommand\algorithmicforeach{\textbf{for each}}
\algdef{S}[FOR]{ForEach}[1]{\algorithmicforeach\ #1\ \algorithmicdo}

\usepackage{hyperref}
\hypersetup{
colorlinks=true,
linkcolor=blue,
filecolor=magenta,
urlcolor=blue,
}

\title{\LARGE \bf
PlaceFormer: Transformer-based Visual Place Recognition using Multi-Scale Patch Selection and Fusion
}
\author{Shyam Sundar Kannan and Byung-Cheol Min
\thanks{The authors are with SMART Lab, Department of Computer and Information Technology, Purdue University, West Lafayette, IN 47907, USA \tt{\small{shyamkannan@purdue.edu | minb@purdue.edu}}}%
}

\begin{document}
\maketitle
\thispagestyle{empty}
\pagestyle{empty}

\begin{abstract}
Visual place recognition is a challenging task in the field of computer vision, and autonomous robotics and vehicles, which aims to identify a location or a place from visual inputs. Contemporary methods in visual place recognition employ convolutional neural networks and utilize every region within the image for the place recognition task. However, the presence of dynamic and distracting elements in the image may impact the effectiveness of the place recognition process. Therefore, it is meaningful to focus on task-relevant regions of the image for improved recognition. In this paper, we present PlaceFormer, a novel transformer-based approach for visual place recognition. PlaceFormer employs patch tokens from the transformer to create global image descriptors, which are then used for image retrieval. To re-rank the retrieved images, PlaceFormer merges the patch tokens from the transformer to form multi-scale patches. Utilizing the transformer's self-attention mechanism, it selects patches that correspond to task-relevant areas in an image. These selected patches undergo geometric verification, generating similarity scores across different patch sizes. Subsequently, spatial scores from each patch size are fused to produce a final similarity score. This score is then used to re-rank the images initially retrieved using global image descriptors. Extensive experiments on benchmark datasets demonstrate that PlaceFormer outperforms several state-of-the-art methods in terms of accuracy and computational efficiency, requiring less time and memory. 
\end{abstract}

\section{Introduction}
\label{sec:intro}
Visual Place Recognition (VPR) is a critical task for localizing autonomous vehicles and robots navigating through dynamic environments, relying on visual input such as images or videos. Visual place recognition is defined as an image retrieval problem \cite{lowry2015visual}, where a query image from an unknown location is compared with a database of reference images from known locations in order to localize the query image. The location of the query image is estimated by identifying the closest matching image in the reference image database. This task is challenging due to variations in seasons, illumination, viewpoint, and occlusions. Typically, two types of image representations are used in VPR tasks: global and patch-level descriptors. Global descriptors \cite{arandjelovic2016netvlad, gordo2017end} provide a succinct image representation in a single vector, facilitating efficient large-scale searches. Patch-level or local descriptors \cite{li2023hot, hausler2021patch, khaliq2019holistic} encode details about specific regions or key points of the image and are used for performing geometric verification between image pairs. 

To enhance performance, VPR is commonly executed in two distinct phases. Initially, a global retrieval is conducted by employing a nearest-neighbor search between the query image's global descriptors and those of the reference images. Subsequently, using the patch descriptors, re-ranking is conducted on the top-$k$ candidate images acquired during the global retrieval process. Re-ranking is typically achieved through cross-matching the patch descriptors of both the query and the reference images, followed by a subsequent geometric verification step. However, the larger size of patch descriptors, which usually encode all regions of an image, can slow down and prolong the re-ranking process. Therefore, it is crucial to extract only the task-relevant regions to facilitate the re-ranking step efficiently.

Contemporary VPR methods rely on Convolutional Neural Networks (CNNs) to extract both global and local image descriptors. In VPR applications, the visual characteristics of a location can undergo significant changes over the long term, including alterations caused by factors such as day-night illumination, falling leaves, and snow. Therefore, a comprehensive grasp of the image's global context is crucial for successful VPR. Nevertheless, CNNs, with their limited receptive fields, are not inherently adept at capturing this global context. Vision Transformers \cite{dosovitskiy2020image}, on the contrary, addresses this CNN limitation by introducing pair-wise attention mechanisms that can capture relationships between any pair of locations within an image. This innovative approach allows the transformer's patch tokens to encode not only local information but also vital global context, enhancing its suitability for VPR tasks. 

\begin{figure}[t]
\centering
\includegraphics[width=0.9\linewidth]{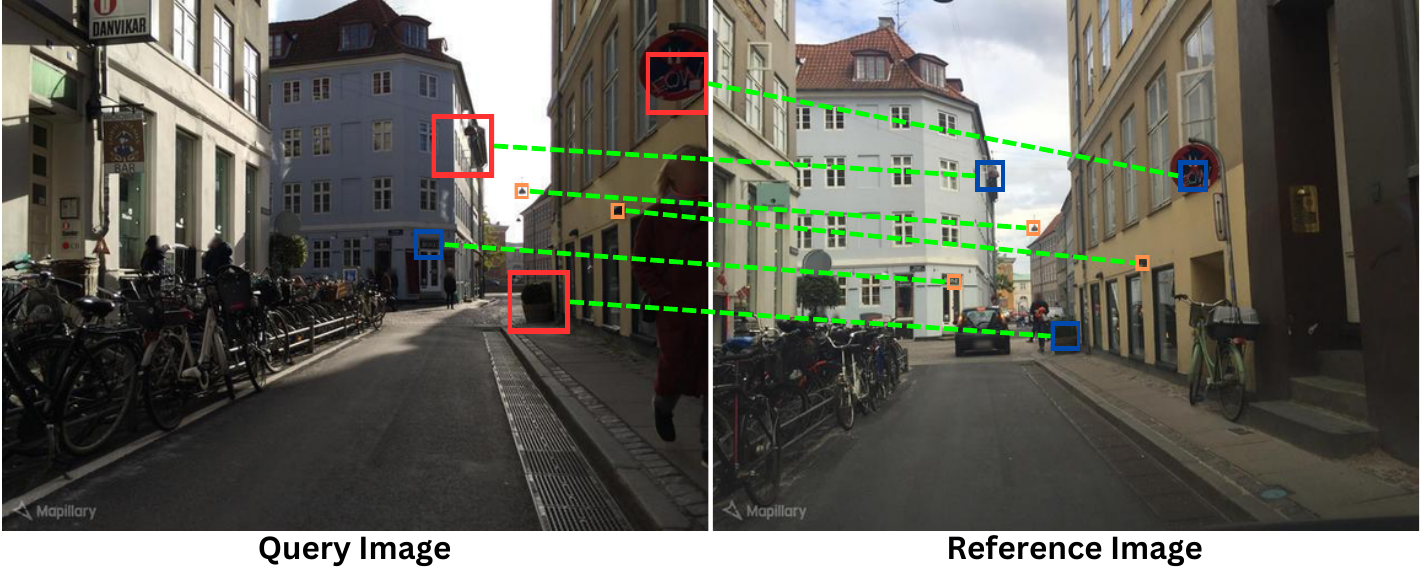}
\vspace{-8pt}

\caption{PlaceFormer leverages patches of varying scales achieved through the fusion of patch tokens in the vision transformer. From these fused patches, key patches (boxes of different colors) are selectively chosen based on the attention scores from the transformer corresponding to that patch. The model then estimates correspondences between key patches of different scales in both the query and reference images which is used for the image retrieval process.} 
\label{fig:intro_pic}
\vspace{-8mm}
\end{figure}

In this paper, we propose \textbf{PlaceFormer}, a novel approach that harnesses the potential of vision transformers to extract robust image representations specifically designed for visual place recognition. The global retrieval process involves aggregating the patch tokens of the transformer and utilizing the same to perform global retrieval. Then, we compute patches of multiple fixed scales on the images by fusing the patch tokens and then, identify key patches amongst them, leveraging the attention map of the vision transformer. \textcolor{black}{Patches of multiple scales are employed to enhance correspondence matching between images despite variations in the scale and viewpoints.}. These key patches pinpoint areas of the image ideal for accurate long-term VPR. Comparing key patches in both query and reference images across different scales, we compute similarity scores using geometric verification. Subsequently, a re-ranking process is carried out based on these similarity scores. In Fig.~\ref{fig:intro_pic}, we illustrate a visual example that highlights key patches of various scales selected based on attention scores (marked with boxes of distinct colors). The green lines represent the correspondences estimated between patches of different scales in both the query and reference images, which are utilized for the re-ranking process.

In summary, the main contributions of our work are:
\begin{itemize}
    \item A vision transformer-based VPR model PlaceFormer that extracts robust global and patch-level image representations.
    \item Attention-based multi-scale patch selection and fusion module that cross-matches patches of different scales and computes a similarity score between an image pair for re-ranking the images. 
    \item Extensive validation of PlaceFormer on numerous VPR benchmarks, and it achieves state-of-the-art performance on several benchmarks while requiring less computation time and memory. 
\end{itemize}

\section{Related Works}
\label{sec:rel_work}

\noindent\textbf{Global Image Descriptors.} The early methodologies for generating global image descriptors initially relied on aggregating local descriptors using techniques such as Bag of Words (BoW) \cite{csurka2004visual} and Vector of Locally Aggregated Descriptors (VLAD) \cite{arandjelovic2013all}. With the advent of deep learning, various methods were developed for aggregating or pooling features obtained through Convolutional Neural Networks, including NetVLAD \cite{arandjelovic2016netvlad}, CRN \cite{jin2017learned}, GeM \cite{radenovic2018fine}, and R-MAC \cite{gordo2017end}. Recently, there have been efforts towards the simultaneous extraction of both global and local descriptors using CNNs  \cite{cao2020unifying}. In the utilization of CNNs for the extraction of global descriptors, the network typically incorporates down-sampling layers to encode task-relevant contextual information. Nevertheless, this downsampling can potentially result in the loss of intricate image details crucial for place recognition. To this end, vision transformer  \cite{dosovitskiy2020image} has been used in \cite{el2021training}, where the \texttt{[class]} tokens from the final transformer layer are employed as global descriptors for image retrieval. Distinct from existing approaches, we leverage a vision transformer to generate global descriptors by pooling the patch tokens from the transformer, facilitating more comprehensive representations of the entire image tailored for intricate visual place recognition tasks. 

\noindent\textbf{Patch-Level Descriptors.} Earlier approaches for extracting patch-level descriptors relied on handcrafted features such as SIFT \cite{lowe1999object}, SURF \cite{bay2008speeded}, and BRIEF \cite{cal2010brief} at key points. However, these features struggled to adapt to the substantial long-term changes typical in place recognition tasks. CNNs have also been employed for patch-level descriptor extraction \cite{chen2018learning, camara2020visual, cao2020unifying, khaliq2019holistic}, capturing features from diverse image regions. Patch-NetVLAD \cite{hausler2021patch} adapted the NetVLAD global descriptor framework to create descriptors for multiple fixed-size patches in an image. Expanding on this idea, Hot-NetVLAD \cite{li2023hot} proposes techniques to identify image patches crucial for place recognition. Focusing on these specific areas optimizes Patch-NetVLAD's application, reducing memory usage and computational demands compared to methods processing all available patches.

TransVPR \cite{wang2022transvpr} combines CNN and vision transformer for place recognition, extracting global and local features by integrating a CNN backbone with transformer layers. Particularly noteworthy is its selection of vital local features for place recognition through the merging of attention maps from diverse transformer layers. $R^2$Former \cite{zhu2023r2former}, another transformer-based method, extends beyond mere feature extraction, employing transformers for both the feature extraction and re-ranking process. These transformer-based methodologies rely on the use of patch tokens from transformers as such to extract patch descriptors. \textcolor{black}{These techniques leverage patch tokens of the same size from vision transformers to estimate correspondences when matching two images. However, variations in viewpoint and scale across images may cause some correspondences to be overlooked, consequently impacting performance.} To overcome this, we propose a local fusion of patch tokens to extract patches of various scales and cross-matching patches of different scales, hence estimating extensive correspondences between the images for the matching process. 


\section{Methodology}
\label{sec:methodology}
\begin{figure*}[h] 
    \centering
    \includegraphics[width=0.84\linewidth]{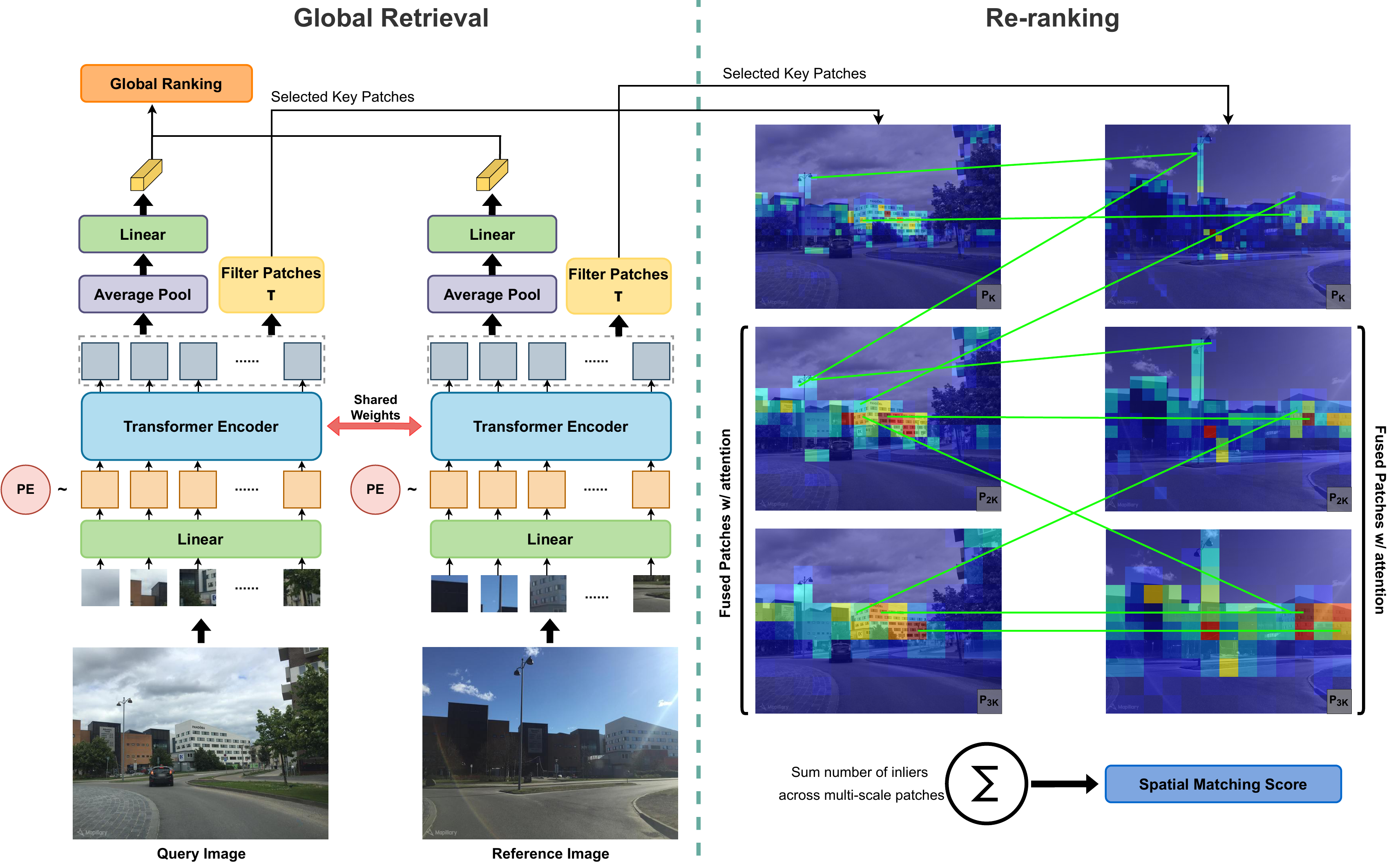}
    \vspace{-10pt}
    \caption{The proposed framework, PlaceFormer, encompasses a two-phase approach for visual place recognition. In the global retrieval phase, patch tokens extracted from the vision transformer undergo pooling and are subsequently processed through a linear layer, resulting in a feature vector utilized for efficient global retrieval. In the re-ranking phase, the patch tokens and attention map from the transformer's last layer are fused to generate patches at multiple scales. Leveraging attention scores, key patches are selectively identified, and correspondences between patches of different scales are computed. In the figure, for brevity, only a few inliers between the patches have been visualized. These inliers contribute to computing a spatial matching score, which is crucial for the re-ranking process. } 
    \label{fig:archi}
    \vspace{-7mm}
\end{figure*}

PlaceFormer employs a vision transformer \cite{dosovitskiy2020image} as its backbone and extracts global descriptors and patch tokens from a given query image. The global descriptors are used to retrieve the top-$k$ candidate images. The patch tokens are fused to create patches of multiple scales and correspondences are computed across patches of different scales. The number of inliers found is used to compute a similarity score which, in turn, is employed to re-rank the candidate images. A depiction of the architecture and functioning of PlaceFormer is provided in Fig.~\ref{fig:archi}.

\subsection{Global Image Retrieval}
\label{sec:global_retrieval}
Given a set of query images $\{I_q\}$ and a corresponding set of reference images $\{I_r\}$, the primary objective of global image retrieval is to create image representations that facilitate the close association of a query image, $I_q$ with a positive reference image, $I_r$ while ensuring a clear distinction from a negative reference image. 

During the training phase, positive samples are identified as reference images that are located within a threshold distance of $10$ meters from the query image. Conversely, negative samples are defined as those reference images that are situated more than $25$ meters away from the query image. This distance threshold is strategically set to ensure that negative samples are distinctly separate from the query image in terms of spatial location, thus providing a clear demarcation between positive and negative associations. The thresholds are set following the common practices from previous works \cite{berton2022deep}. 

An input image $I \in \mathbb{R}^{h \times w \times c}$ is given as input to the transformer block, where $h, w, c$ are the height, width, and number of channels in the image. To extract the descriptors from the image, we use a standard vision transformer. The image is divided into patches of size $p \times p$. Each patch is then transformed into patch tokens $P \in \mathbb{R}^{n \times d}$ through linear projection, where the number of patch tokens, $n = h/p \times w/p$ and $d$ is the dimension of each token. Learnable positional embedding $PE \in \mathbb{R}^{(n+1) \times d}$ are added to the tokens. These positional embeddings add positional information about the position of each token within the image. In cases, of varying sizes of the input image, the positional embedding is interpolated to the size of the input image. 

Within the transformer blocks, Multi-Head Attention (MHA) is used. In other words, each transformer block has multiple heads that compute pairwise attention values. Each attention head projects the inputs given to it into query $Q$, key $K$, and value $V$ each with dimension $d$. Now the basic attention at each head is computed as
\begin{equation}
    \centering
    Attention (Q, K, V) = softmax(\frac{Q.K^T}{\sqrt{d_k}}.V).
\end{equation}

These attention values are used later in the re-ranking phase of PlaceFormer. The vision transformer outputs $n$ patch tokens, $P_L$ with each of size $d$. This output needs to be compressed into a concise vector such that quick global retrieval of images can be performed with this vector. First, the $n$ patch tokens undergo compression through an average pooling operation as:
\begin{equation}
    \centering
    M = \frac{1}{n} \sum_n P_L.
\end{equation}

The average pooling produces a linear vector, $M$, with length $d$. To further reduce the vector's size, a linear layer is applied, resulting in the generation of the global feature descriptor $G$ with a dimensionality of 256. Subsequently, image retrieval is executed between the images in sets ${I_q}$ and ${I_r}$ through nearest neighbor search utilizing Euclidean distance. In accordance with the distance metric, for each query image in ${I_q}$, the top-$k$ closest reference images from ${I_r}$ are retrieved, and these are subjected to re-ranking in the subsequent step.

\noindent{\textbf{Loss function.}}  During the training phase, the model is optimized using Triplet margin loss, aiming to minimize the distance between pairs of positive images and concurrently maximize the distance between pairs of negative images. Let $G_q$, $G_p$, and $G_n$ be the global feature descriptors for the query image $I_q$, positive reference sample $I_q$, and negative reference sample $I_n$. Now triple loss is computed as:
\begin{equation}
    \centering
    L = max(||G_q-G_p||^2-||G_q-G_n||^2+m,0)
\end{equation}
where $m$ is the margin, a hyperparameter used in the training process. In Fig.~\ref{fig:archi}, the left segment illustrates the procedure of extracting global features for global retrieval by employing a query and a reference image. Throughout the training process, the same architecture is used to extract features from all image samples.

\subsection{Patch Token Fusion}
\label{sec:patch_fusion}
In theory, patch tokens from any layer of the transformer can serve as patch-level descriptors. However, in PlaceFormer, we specifically utilize the patch tokens extracted from the last layer of the transformer, $P_L$ to generate the patch descriptors. $P_L$ comprises patch tokens, with each token corresponding to a patch of size $p \times p$ in the image. To generate patch tokens corresponding to larger regions in the image, average pooling is performed on $P_L$ using kernels of sizes $2$ and $3$. This operation results in fused patch tokens, $P_{L2}$ and $P_{L3}$, where each token now corresponds to a patch of size $2p \times 2p$ and $3p \times 3p$ on the image, respectively.

Similarly, the attention map from the last layer of the transformer, $A_L$, undergoes average pooling with kernel sizes $2$ and $3$ to generate fused attention maps, $A_{2L}$ and $A_{3L}$, corresponding to the fused patch tokens $P_{L2}$ and $P_{L3}$.

\subsection{Attention-based Key Patch Selection}
\label{sec:patch_selection}
To optimize the re-ranking of candidate images and minimize computational overhead, we concentrate on key patch tokens identified through the attention map of the transformer. This strategy emphasizes the most pertinent patches, improving efficiency without compromising accuracy. Utilizing the attention maps $A_L$, $A_{2L}$, and $A_{3L}$, we choose the top $400$, $200$, and $50$ patches\footnote{\textcolor{black}{The limit on the number of patches was determined through experimentation in the development phase. Based on the attention score threshold, $\tau$, the chosen number of patches consistently ranged around 400, 200, and 50 for each respective patch size. These findings led to capping the maximum number of patches to establish an upper limit on memory requirements. It was also found that increasing these limits did not lead to any significant increase in performance.}}, respectively, from $P_{L}$, $P_{L2}$, and $P_{L3}$ based on their attention scores. Subsequently, we refine the selection by filtering the chosen patch tokens, retaining only those with an attention score surpassing a predefined threshold $\tau$, consequently identifying the key patches across the three scales $P_{K}$, $P_{K2}$, and $P_{K3}$, where $P_{K} \subseteq P_{L}$, $P_{K2} \subseteq P_{L2}$ and $P_{K3} \subseteq P_{L3}$.

\subsection{Mutual Nearest Neighbors}
\label{sec:nearest_neighbors}
Considering a set of selected key patch descriptors for a query and reference image as $\{p_{i}^{q}\}_{i=1}^{n_q}$ and $\{p_{i}^{r}\}_{i=1}^{n_r}$, where $n_q$ and $n_r$ denote the total number of key patches in the query and the reference images, we derive descriptor pairs via mutual nearest neighbor by exhaustively comparing the two descriptor sets. The set of mutual nearest neighbors, $\mathbb{P}$ is computed as:
\begin{equation}
    \centering
    \mathbb{P}=\left\{(i, j): i={N N}_{r}\left({p}_{j}^{q}\right), j={N N}_{q}\left({p}_{i}^{r}\right)\right\}
\end{equation}

where $\mathrm{NN}_{q}({p})=\operatorname{argmin}_{j}\left\|{p}-{p}_{j}^{q}\right\|_{2}$ and $\mathrm{NN}_{r}({p})=\operatorname{argmin}_{i}\left\|{p}-{p}_{i}^{r}\right\|_{2}$ computes the nearest neighbor matches between query and the reference-based on Euclidean distance. Utilizing the set of matching patches, a spatial matching score can be computed by assessing the number of inliers obtained during the fitting of the homography through RANSAC, based on the corresponding patches. When fitting the homography, we assume that each patch corresponds to a 2D image point with coordinates at the center of the patch. For homography fitting, the tolerance error for inliers is set at 1.5 times the patch size.

\subsection{Multi-Scale Patch Matching}
\label{sec:patch_matching}
Now, we employ the spatial matching score formulation to compare key patches of different scales. In PlaceFormer, the computation of the spatial matching score is performed across three combinations of key patch scales. First, we calculate $s_{1,1}$, representing the spatial matching scores between non-fused key patch tokens $P_K$ from the query and reference images. Subsequently, we compute $s_{1,2}$, denoting the spatial matching score between the combination of $P_K$ and $P_{2K}$ from the query and reference images. Finally, we estimate $s_{2,3}$, the spatial matching score between the combination of $P_{2K}$ and $P_{3K}$ from the query and reference images.

The final spatial score $S_{spatial}$\footnote{\textcolor{black}{Weights were introduced during development to scale individual spatial matching scores, yet no significant changes in the results were observed. Consequently, these weights were omitted from the final method.}} is computed by the summation of the spatial matching scores computed across patches of different scales, 
\begin{equation}
    \centering
    \begin{array}{r}
        S_{spatial} = s_{1,1} + s_{1,2} + s_{2,3}.
    \end{array}
\end{equation}

Once the $S_{spatial}$ is computed for all the $k$ candidate reference images extracted through global retrieval, the images are re-ranked and the final list of matching reference images is estimated.


\section{Experiments}
\label{sec:expt}
\subsection{Implementation and Training Details}
\label{sec:implementation}
PlaceFormer is developed using the PyTorch framework. The training and testing of the models are performed on an NVIDIA RTX 3090 graphics card. For the base encoder, we utilize the Vision Transformer Small (ViT-S) model \cite{dosovitskiy2020image}. This model is characterized by its $12$ layers of transformers, each containing $12$ heads. The transformer within this model is designed to extract patch tokens, each with a dimension ($d$) of $384$. We chose patch size, $p$ of $16\times16$ aligning with the model's architecture. To maintain consistency with previous research and ensure compatibility, all images used in both training and testing phases are resized to a resolution of $640 \times 480$. A key patch filtering threshold, $\tau$ of $0.01$ is used in the implementation. The re-ranking is performed on top-100 ($k=100$) candidates retrieved through global retrieval. A margin, $m$ of $0.01$ is used in the triplet loss. 

\noindent\textbf{Training.} The transformer is initialized with pre-trained weights on ImageNet-1K for the training process. The model is trained using the MSLS train dataset \cite{warburg2020mapillary}. MSLS is chosen as the train set due to its diversity in scene types and the presence of various environmental variations, providing a comprehensive training environment. The model is further fine-tuned using Pittsburgh 30K (Pitts30K) training set \cite{teichmann2019detect}. Both the positive and negative samples are pre-computed before training to optimize the training duration. The training utilized the Adam optimizer, in conjunction with a cosine learning rate scheduler. The initial learning rate is set at $0.0005$. The training is continued until there is no further improvement in accuracy on the validation set.

\begin{table*}
\centering
\caption{Comparison of PlaceFormer with state-of-the-art methods on benchmark datasets. }
\vspace{-2mm}
\label{tab:qual_resu}
\resizebox{\linewidth}{!}{%
\begin{tblr}{
  column{2} = {c},
  column{3} = {c},
  column{4} = {c},
  column{6} = {c},
  column{7} = {c},
  column{8} = {c},
  column{14} = {c},
  column{15} = {c},
  column{16} = {c},
  column{18} = {c},
  column{19} = {c},
  column{20} = {c},
  cell{1}{1} = {r=2}{c},
  cell{1}{2} = {c=3}{},
  cell{1}{5} = {c},
  cell{1}{6} = {c=3}{},
  cell{1}{10} = {c=3}{c},
  cell{1}{14} = {c=3}{},
  cell{1}{17} = {c},
  cell{1}{18} = {c=3}{},
  cell{1}{22} = {c=3}{c,fg=black},
  cell{2}{5} = {c},
  cell{2}{17} = {c},
  cell{2}{22} = {fg=black},
  cell{2}{23} = {fg=black},
  cell{2}{24} = {fg=black},
  cell{3}{5} = {c},
  cell{3}{10} = {c},
  cell{3}{11} = {c},
  cell{3}{12} = {c},
  cell{3}{17} = {c},
  cell{3}{22} = {c,fg=black},
  cell{3}{23} = {c,fg=black},
  cell{3}{24} = {c,fg=black},
  cell{4}{5} = {c},
  cell{4}{10} = {c},
  cell{4}{11} = {c},
  cell{4}{12} = {c},
  cell{4}{17} = {c},
  cell{4}{22} = {c,fg=black},
  cell{4}{23} = {c,fg=black},
  cell{4}{24} = {c,fg=black},
  cell{5}{1} = {fg=black},
  cell{5}{2} = {fg=black},
  cell{5}{3} = {fg=black},
  cell{5}{4} = {fg=black},
  cell{5}{6} = {fg=black},
  cell{5}{7} = {fg=black},
  cell{5}{8} = {fg=black},
  cell{5}{10} = {c,fg=black},
  cell{5}{11} = {c,fg=black},
  cell{5}{12} = {c,fg=black},
  cell{5}{14} = {fg=black},
  cell{5}{15} = {fg=black},
  cell{5}{16} = {fg=black},
  cell{5}{18} = {fg=black},
  cell{5}{19} = {fg=black},
  cell{5}{20} = {fg=black},
  cell{5}{22} = {c,fg=black},
  cell{5}{23} = {c,fg=black},
  cell{5}{24} = {c,fg=black},
  cell{6}{1} = {fg=black},
  cell{6}{2} = {fg=black},
  cell{6}{3} = {fg=black},
  cell{6}{4} = {fg=black},
  cell{6}{5} = {c},
  cell{6}{6} = {fg=black},
  cell{6}{7} = {fg=black},
  cell{6}{8} = {fg=black},
  cell{6}{10} = {c,fg=black},
  cell{6}{11} = {c,fg=black},
  cell{6}{12} = {c,fg=black},
  cell{6}{14} = {fg=black},
  cell{6}{15} = {fg=black},
  cell{6}{16} = {fg=black},
  cell{6}{17} = {c},
  cell{6}{18} = {fg=black},
  cell{6}{19} = {fg=black},
  cell{6}{20} = {fg=black},
  cell{6}{22} = {c,fg=black},
  cell{6}{23} = {c,fg=black},
  cell{6}{24} = {c,fg=black},
  cell{7}{5} = {c},
  cell{7}{10} = {c},
  cell{7}{11} = {c},
  cell{7}{12} = {c},
  cell{7}{17} = {c},
  cell{7}{22} = {c,fg=black},
  cell{7}{23} = {c,fg=black},
  cell{7}{24} = {c,fg=black},
  cell{8}{5} = {c},
  cell{8}{10} = {c},
  cell{8}{11} = {c},
  cell{8}{12} = {c},
  cell{8}{17} = {c},
  cell{8}{22} = {c,fg=black},
  cell{8}{23} = {c,fg=black},
  cell{8}{24} = {c,fg=black},
  cell{9}{10} = {c},
  cell{9}{11} = {c},
  cell{9}{12} = {c},
  cell{9}{22} = {c,fg=black},
  cell{9}{23} = {c,fg=black},
  cell{9}{24} = {c,fg=black},
  cell{10}{5} = {c},
  cell{10}{10} = {c},
  cell{10}{11} = {c},
  cell{10}{12} = {c},
  cell{10}{17} = {c},
  cell{10}{22} = {c,fg=black},
  cell{10}{23} = {c,fg=black},
  cell{10}{24} = {c,fg=black},
  cell{11}{5} = {c},
  cell{11}{10} = {c},
  cell{11}{11} = {c},
  cell{11}{12} = {c},
  cell{11}{17} = {c},
  cell{11}{22} = {c,fg=black},
  cell{11}{23} = {c,fg=black},
  cell{11}{24} = {c,fg=black},
  cell{12}{22} = {c,fg=black},
  cell{12}{23} = {c,fg=black},
  cell{12}{24} = {c,fg=black},
  cell{13}{22} = {c,fg=black},
  cell{13}{23} = {c,fg=black},
  cell{13}{24} = {c,fg=black},
  hline{1,3,8,13-14} = {-}{},
  hline{2} = {2-16,18-24}{},
}
Method                       & MSLS Validation \cite{warburg2020mapillary} &               &               &  & MSLS Challenge \cite{warburg2020mapillary} &               &               &  & Nordland \cite{sunderhauf2013we}      &               &               &  & Pitts30K \cite{teichmann2019detect}      &               &               &  & Tokyo 24/7 \cite{torii201524}    &               &               &  & Robotcar-S2 \cite{sattler2018benchmarking} &           &             \\
                             & R@1             & R@5           & R@10          &  & R@1            & R@5           & R@10          &  & R@1           & R@5           & R@10          &  & R@1           & R@5           & R@10          &  & R@1           & R@5           & R@10          &  & .25m/2$^o$  & .5m/5$^o$ & 5.0m/10$^o$ \\
NetVLAD \cite{arandjelovic2016netvlad}    & 60.8            & 74.3          & 79.5          &  & 35.1           & 47.4          & 51.7          &  & 13.6          & 21.4          & 25.2          &  & 81.9          & 91.2          & 93.7          &  & 64.8          & 78.4          & 81.6          &  & 5.6         & 20.7      & 71.8        \\
SFRS \cite{ge2020self}                         & 69.2            & 80.3          & 83.1          &  & 41.5           & 52.0          & 56.3          &  & 16.4          & 26.3          & 29.7          &  & 89.4          & 94.7          & 95.9          &  & 85.4          & 91.1          & 93.3          &  & 8.0         & 27.3      & 80.4        \\
CosPlace \cite{berton2022rethinking}                     & 85.2            & 92.3          & 93.2          &  & 60.9           & 71.7          & 76.7          &  & 54.7          & 70.9          & 77.9          &  & 89.0          & 94.7          & 96.1          &  & 81.0          & 90.8          & 93.7          &  &   8.2          &  29.9         &   83.7          \\
MixVPR \cite{ali2023mixvpr}                      & 88.0            & 92.7          & 94.6          &  & 64.0           & 75.9          & 80.6          &  & 58.4          & 74.6          & 80.0          &  & 91.9          & 95.9          & 96.7          &  & 85.1          & 91.7          & \textbf{94.3} &  &  8.9           &   33.3        &      86.5       \\
PlaceFormer (w/o re-ranking) & 80.0            & 90.0          & 93.0          &  & 58.0           & 76.7          & 81.6          &  & 26.1          & 40.3          & 47.6          &  & 82.7          & 93.0          & 95.1          &  & 57.5          & 74.6          & 80.3          &  & 3.2         & 15.7      & 60.3        \\
SP-SuperGlue \cite{sarlin2020superglue}                 & 78.1            & 81.9          & 84.3          &  & 50.6           & 56.9          & 58.3          &  & 29.1          & 33.5          & 34.3          &  & 87.2          & 94.8          & 96.4          &  & 88.2          & 90.2          & 90.2          &  & 9.5         & 35.4      & 85.4        \\
DELG \cite{cao2020unifying}                         & 83.2            & 89.3          & 91.1          &  & 52.2           & 61.9          & 65.4          &  & 51.0          & 63.9          & 66.7          &  & 89.8          & 95.3          & 96.7          &  & 86.4          & \textbf{92.4} & 93.0          &  & 2.2         & 8.4       & 76.8        \\
Patch-NetVLAD \cite{hausler2021patch}                & 79.5            & 86.2          & 87.7          &  & 48.1           & 57.6          & 60.5          &  & 46.4          & 58.0          & 60.4          &  & 88.7          & 94.5          & 95.9          &  & 86.0          & 88.6          & 90.5          &  & 9.6         & 35.3      & 90.9        \\
TransVPR \cite{wang2022transvpr}                     & 86.8            & 91.2          & 92.4          &  & 63.9           & 74.0          & 77.5          &  & 58.8          & \textbf{75.0} & \textbf{78.7} &  & 89.0          & 94.9          & 96.2          &  & 79.0          & 82.2          & 85.1          &  & 9.8         & 34.7      & 80.0        \\
$R^2$Former \cite{zhu2023r2former}                  & 89.7            & \textbf{95.0} & \textbf{96.2} &  & \textbf{73.0 } & \textbf{85.9} & \textbf{88.8} &  & 60.6          & 66.8          & 68.7          &  & 91.1          & 95.2          & 96.3          &  & \textbf{88.6} & 91.4          & 91.7          &  &  10.5            &   35.2        &   85.2          \\
PlaceFormer (Ours)           & \textbf{89.9}   & 94.3          & 95.4          &  & 71.9           & 85.4          & 88.7          &  & \textbf{65.3} & 70.5          & 72.4          &  & \textbf{92.4} & \textbf{96.5} & \textbf{97.4} &  & 87.6          & 89.2          & 91.5          &  & \textbf{10.8}        & \textbf{37.6}      & \textbf{92.1}        
\end{tblr}
}
\vspace{-10pt}
\end{table*}

\subsection{Datasets}
\label{sec:dataset}
We evaluate PlaceFormer on multiple public benchmark datasets. These included the Nordland \cite{sunderhauf2013we}, Pittsburgh 30K (Pitts30K) \cite{teichmann2019detect}, Tokyo24/7 \cite{torii201524}, \textcolor{black}{RobotCar Seasons v2 (RobotCar-S2)} \cite{sattler2018benchmarking} and Mapillary Street-Level Sequences (MSLS) \cite{warburg2020mapillary}. Each of these datasets offers a unique set of challenges, encompassing a wide array of environments and conditions that are crucial for thorough performance assessment. All images used in the evaluation are resized to a uniform resolution of $640 \times 480$ to be consistent with other methods. The model trained on MSLS is used to evaluate MSLS and Nordland datasets. The model fine-tuned on Pitts30K is used to evaluate Pitts30K, \textcolor{black}{RobotCar-S2}, and Tokyo 24/7 (urban scenarios). 

\subsection{Metrics}
\label{sec:metrics}
For MSLS, Nordland, and Pitta30K datasets, Recall@K is used as the primary metric for evaluation. This metric quantifies the percentage of query images that are correctly localized within a dataset. It does so by determining whether at least one of the top K-ranked reference images falls within a specified threshold distance from the query image. For our evaluation, we followed the precedent set in prior works \cite{berton2022deep, wang2022transvpr, hausler2021patch}, using a threshold distance of $25$ meters. The Recall@K is measured for K values of 1, 5, and 10. \textcolor{black}{For the RobotCar-S2 dataset, we use the pose of the closest matching image as the estimated pose and compute recall under three default error thresholds.} 

\subsection{Comparison with State-of-the-arts}
\label{sec:comp}
In our comparative analysis, PlaceFormer is benchmarked against a range of state-of-the-art methods to demonstrate its efficacy in visual place recognition. We evaluated it alongside methods such as \textcolor{black}{NetVLAD \cite{arandjelovic2016netvlad}, SFRS \cite{ge2020self}, CosPlace \cite{berton2022rethinking}, and MixVPR \cite{ali2023mixvpr}}, which primarily utilize global image representations. Alongside these methods, we also presented the results of our global retrieval to offer a comprehensive comparison. Furthermore, we compared PlaceFormer with techniques that employ both global and local features for retrieval and ranking. This included comparisons with Patch-NetVLAD \cite{hausler2021patch} and DELG \cite{cao2020unifying}. 

In addition, we included a comparison with a high-performing baseline, SP-SuperGlue \cite{sarlin2020superglue} which combines NetVLAD for retrieval and SuperGlue for matching patch-level descriptors. Lastly, our analysis also encompassed comparisons with TranVPR \cite{wang2022transvpr} and $R^2$Former \cite{zhu2023r2former}, which are prominent in utilizing transformers for feature extraction in VPR. 

\vspace{-8pt}
\section{Results}
\label{sec:result}
\subsection{Quantitative Results}
\label{sec:quan_res}
\textcolor{black}{The quantitative performance of PlaceFormer, in comparison to other approaches, is detailed in Table \ref{tab:qual_resu}. PlaceFormer without local re-ranking outperforms traditional global retrieval methods like NetVLAD and SFRS in MSLS validation, MSLS challenge, and Nordland datasets, and yields comparable performance in Pitts30K and Tokyo 24/7 datasets. Furthermore, PlaceFormer without re-ranking outperforms multiple re-ranking methods as well, making it suitable to be used even without re-ranking based on specific requirements.}

PlaceFormer with re-ranking achieves competitive results on all datasets. It outperforms all comparable methods on MSLS validation, Nordland, and Pitts30K in Recall@1 with absolute differences of $0.2\%$, $5.3\%$, and $1.3\%$ compared to the best-performing baseline of $R^2$Former. \textcolor{black}{When computing the average performance across all datasets, PlaceFormer demonstrates a substantial superiority over competing methods, surpassing NetVLAD, SFRS, CosPLace, MixVPR, SP-SuperGlue, DELG, PatchNetVLAD, TransVPR, and $R^2$Former by margins of $30.8\%$, $21.04\%$, $11.6\%$, $3.94\%$, $14.78\%$, $8.9\%$, $11.68\%$, $5.92\%$, and $0.82\%$ in Recall@1 scores, respectively.} While PlaceFormer generally surpasses other state-of-the-art methods, it falls short of outperforming $R^2$Former on certain datasets such as MSLS Challenge and Tokyo 24/7. It is important to highlight that our method is specifically trained for global retrieval only, whereas $R^2$Former undergoes training for re-ranking, contributing to its enhanced performance. \textcolor{black}{PlaceFormer demonstrates superior performance compared to other methods across all three thresholds in the RobotCar S-2 dataset. Notably, it excels particularly under the $5.0m/10^o$ threshold, attributed to its adeptness in managing viewpoint discrepancies through multi-scale patch matching, distinguishing itself significantly from alternative approaches.}

\subsection{Latency and Memory}
\label{sec:latency}
Table \ref{tab:lat_memory} shows the computational time and memory requirements for the state-of-the-art methods and PlaceFormer for a single query image.  SP-SuperGlue, DELG, Patch-NetVLAD, and TransVPR all adopt RANSAC for their matching process, a strategy similar to PlaceFormer. However, PlaceFormer exhibits superior efficiency, being $18.4$, $12.3$, $148.4$, and $5$ times faster in terms of extraction latency, and $7.11$, $32.76$, $6.95$, and $2.9$ times faster in terms of matching latency compared to these methods. The enhanced speed of PlaceFormer in matching can be attributed to its approach of selecting key patches using attention scores, resulting in a more focused set of points for which the homography needs to be computed during RANSAC. This optimization leads to a significant reduction in computation time. \textcolor{black}{MixVPR though requires the least extraction time, the method extracts a vector of size $4096$ which makes the global retrieval a tedious task}. $R^2$Former requires a similar extraction time as that of PlaceFormer, but it needs less time for matching due to the use of transformer blocks for the matching process. 

\begin{table}[t]
\centering
\caption{Time taken for extracting features, matching descriptors, and memory requirements for the descriptor for state-of-the-art methods and PlaceFormer. For global retrieval methods, matching time and memory requirements are negligible and hence denoted with a `--'.}
\vspace{-2mm}
\label{tab:lat_memory}
\begin{tblr}{
  row{1} = {c},
  cell{2}{2} = {c},
  cell{2}{3} = {c},
  cell{2}{4} = {c},
  cell{3}{2} = {c},
  cell{3}{3} = {c},
  cell{3}{4} = {c},
  cell{4}{2} = {c},
  cell{4}{3} = {c},
  cell{4}{4} = {c},
  cell{5}{2} = {c},
  cell{5}{3} = {c},
  cell{5}{4} = {c},
  cell{6}{2} = {c},
  cell{6}{3} = {c},
  cell{6}{4} = {c},
  cell{7}{2} = {c},
  cell{7}{3} = {c},
  cell{7}{4} = {c},
  cell{8}{2} = {c},
  cell{8}{3} = {c},
  cell{8}{4} = {c},
  cell{9}{2} = {c},
  cell{9}{3} = {c},
  cell{9}{4} = {c},
  cell{10}{2} = {c},
  cell{10}{3} = {c},
  cell{10}{4} = {c},
  hline{1-2,10-11} = {-}{},
}
{\\Method}        & {Extraction\\Latency (ms)} & {Matching\\Latency (s)} & {Memory\\(MB)} \\
NetVLAD \cite{arandjelovic2016netvlad}      & 17                         & --                       & --              \\
SFRS \cite{ge2020self}          & 203                        & --                       & --              \\
\textcolor{black}{MixVPR} \cite{ali2023mixvpr}      & \textcolor{black}{6}             & --            & --              \\
SP-SuperGlue \cite{sarlin2020superglue}  & 166                        & 7.83                    & 1.93           \\
DELG \cite{cao2020unifying}          & 197                        & 36.04                   & 0.37           \\
Patch-NetVLAD \cite{hausler2021patch} & 1336                       & 7.65                    & 44.14          \\
TransVPR \cite{wang2022transvpr}      & 45                         & 3.19                    & 1.17           \\
$R^2$Former \cite{zhu2023r2former}      & 9                          & 0.3                       & 0.5              \\
PlaceFormer (Ours)   & 9                          & 1.1                       & 1.07              
\end{tblr}
\vspace{-7mm}
\end{table}

PlaceFormer exhibits a comparable memory footprint to SP-SuperGlue and TransVPR, all consuming approximately $1.07$ MB. In contrast, PatchNet-VLAD requires significantly more memory due to the necessity of storing patches across various scales. Notably, PlaceFormer optimizes memory usage by storing fewer patches as the patch size increases. While DELG and $R^2$Former have a smaller memory footprint than PlaceFormer, it is important to note trade-offs. DELG, while efficient in memory consumption, requires a substantial amount of time for the matching process. On the other hand, $R^2$Former, while also having lower memory requirements, involves a two-stage training process to compress features, which may be considered a more intricate and time-consuming task.


\subsection{Ablation Study}
We perform multiple ablation experiments to further affirm the design choices made in PlaceFormer.

\noindent{\textbf{Patch Sizes.}}
To assess the efficacy of employing fused patches and explore various combinations, we conduct ablations on different patch sizes and their amalgamations. In Table \ref{tab:abl_patch}, we present the performance of PlaceFormer using patches of varying sizes and combinations for re-ranking. Here, $P_K$ represents key patches selected from the transformer's patch tokens. $P_{2K}$ and $P_{3K}$ denote fused key patches obtained through average pooling with kernel sizes of $2$ and $3$. When using patches of different sizes independently for re-ranking, non-fused patches $P_K$ yield the best results. This outcome can be attributed to the abundance of key patches and their correspondence to smaller regions, facilitating precise homography estimation.

The utilization of patches at multiple scales during correspondence estimation resulted in improved performance, contingent on the combination of patches employed. Specifically, combining non-fused key patches $P_K$ with the first level of fused patches $P_{2K}$, denoted as $P_K$ \& $P_{2K}$ in Table \ref{tab:abl_patch}, yielded increased recall values compared to using patches of the same size independently. This suggests that the synergistic use of key patches at different scales contributes positively to the overall performance, highlighting the effectiveness of incorporating multi-scale information for correspondence estimation in PlaceFormer. 

\begin{figure*}[h] 
    \centering
    \vspace{-3mm}
    \includegraphics[width=0.84\linewidth]{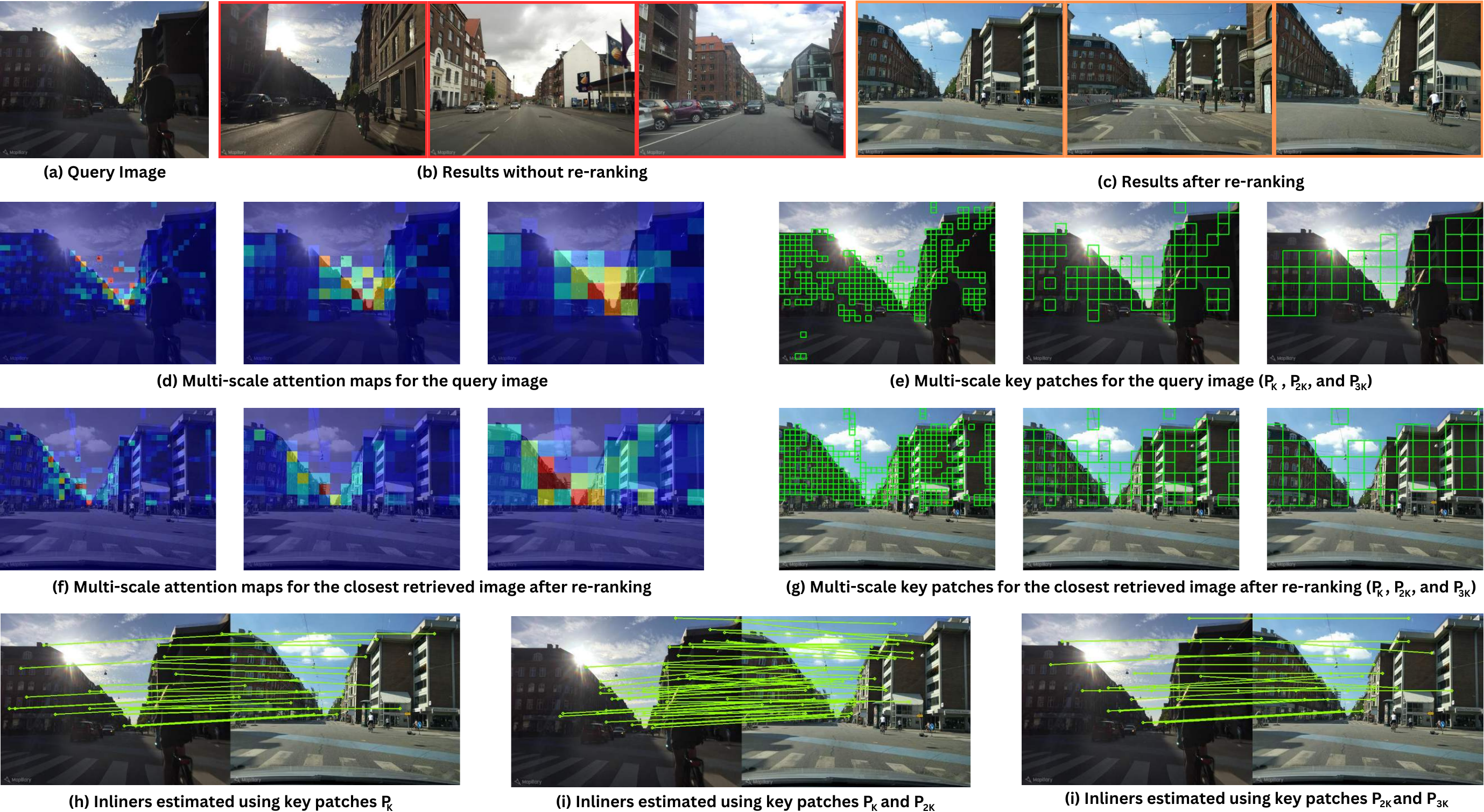}
    \vspace{-5pt}
    \caption{Visualization of global retrieval and re-ranking results (\textcolor{red}{red} box- incorrect retrieval and \textcolor{orange}{orange} box - correct retrieval); the attention maps and key patches for query and closest retrieved image at multiple scales; and the inliers estimated across patches of various scales.} 
    \label{fig:viz_results}
    \vspace{-5mm}
\end{figure*}

\begin{table}[t]
\centering
\caption{Ablation study on various patch sizes and their combination on MSLS validation.}
\label{tab:abl_patch}
\vspace{-2mm}
\begin{tblr}{
  width = \linewidth,
  colspec = {Q[702]Q[73]Q[73]Q[88]},
  cell{1}{2} = {c},
  cell{1}{3} = {c},
  cell{1}{4} = {c},
  cell{2}{2} = {c},
  cell{2}{3} = {c},
  cell{2}{4} = {c},
  cell{3}{2} = {c},
  cell{3}{3} = {c},
  cell{3}{4} = {c},
  cell{4}{2} = {c},
  cell{4}{3} = {c},
  cell{4}{4} = {c},
  cell{5}{2} = {c},
  cell{5}{3} = {c},
  cell{5}{4} = {c},
  cell{6}{2} = {c},
  cell{6}{3} = {c},
  cell{6}{4} = {c},
  cell{7}{2} = {c},
  cell{7}{3} = {c},
  cell{7}{4} = {c},
  cell{9}{2} = {c},
  cell{9}{3} = {c},
  cell{9}{4} = {c},
  cell{10}{2} = {c},
  cell{10}{3} = {c},
  cell{10}{4} = {c},
  cell{11}{2} = {c},
  cell{11}{3} = {c},
  cell{11}{4} = {c},
  hline{1-2,12} = {-}{},
}
Patch Size                                                      & R@1  & R@5  & R@10 \\
$P_K$                                                           & 87.3 & 93.4 & 94.6 \\
$P_{2K}$                                                        & 86.9 & 92.7 & 93.1 \\
$P_{3K}$                                                        & 84.3 & 91.8 & 92.4 \\
$P_K$ \& $P_{2K}$                                               & 88.4 & 93.9 & 94.9 \\
$P_K$ \& $P_{3K}$                                               & 88.1 & 93.5 & 94.7 \\
$P_K$ \& $P_{2K}$ \& $P_{3K}$                                   & 85.4 & 92.7 & 93.1 \\
$P_K$ + $P_K$ \& $P_{2K}$                                       & 88.9 & 93.8 & 95.1 \\
{$P_K$~+~$P_K$~\&~$P_{2K}$~+~$P_{2K}$~\&~$P_{3K}$}   & \textbf{89.9} & \textbf{94.3} & \textbf{95.4} \\
$P_K$ + $P_{K}$ \& $P_{2K}$ + $P_K$ \& $P_{2K}$ \& $P_{3K}$       & 87.9 & 93.5 & 94.9 \\
$P_K$ + $P_{K}$ \& $P_{3K}$ + $P_K$ \& $P_{2K}$ \& $P_{3K}$    & 86.4 & 91.4 & 93.3
\end{tblr}
\vspace{-5mm}
\end{table}

Further, we aggregated the number of inliers estimated through different combinations of patch sizes to assess which combination yields the most effective results for re-ranking. Through experiments, we found that utilizing the sum of inliers estimated using non-fused key patches $P_K$; non-fused key patches $P_K$ with the first level of fused patches $P_{2K}$ ($P_K$ \& $P_{2K}$); and non-fused key patches $P_K$ with the second level of fused patches $P_{3K}$ ($P_K$ \& $P_{3K}$), denoted as $P_K$ + $P_K$ \& $P_{2K}$ + $P_K$ \& $P_{3K}$ in Table \ref{tab:abl_patch} gave the best recall values. This combination of key patches is set as default and used for all the experiments. 

Furthermore, the simultaneous utilization of patches of all three sizes ($P_K$ \& $P_{2K}$ \& $P_{3K}$) generally resulted in a decrease in recall values. This suggests that combining patches of more than two sizes for correspondence estimation may not be well-suited for optimal performance. Additionally, patches with a size exceeding that of $P_{3K}$ are not considered, as such patches may correspond to a significantly large area in the image and may not be optimal for homography computation.

\begin{table}[h]
\centering
\caption{Ablation study on the threshold for patch selection on MSLS validation.}
\label{tab:threshold}
\vspace{-2mm}
\begin{tblr}{
  width = \linewidth,
  colspec = {Q[310]Q[173]Q[173]Q[210]},
  column{even} = {c},
  column{3} = {c},
  hline{1-2,11} = {-}{},
}
Threshold, $\tau$ & R@1  & R@5  & R@10 \\
0.008     & 85.3 & 91.2 & 92.1 \\
0.007     & 86.9 & 92.7 & 93.1 \\
0.009     & 88.7 & 93.2 & 95.1 \\
\textbf{0.01}      & \textbf{89.9} & \textbf{94.3} & \textbf{95.4} \\
0.011     & 89.1 & 93.9 & 94.8 \\
0.012     & 88.2 & 93.1 & 93.8 \\
0.015     & 86.4 & 91.4 & 92.5 \\
0.02      & 86.2 & 91.2 & 92.0 \\
0.05      & 84.2 & 90.9 & 91.9 
\end{tblr}
\vspace{-7pt}
\end{table}

\noindent{\textbf{Key Patch Selection Threshold.}}
We eliminate patch tokens and fused patches associated with potentially non-informative image regions, excluding them from the correspondence estimation. Employing a threshold parameter, $\tau$, only patches with an attention score exceeding $\tau$ contribute to the correspondence estimation process. In Table \ref{tab:threshold}, we present ablation results, assessing the impact of different $\tau$ values on MSLS validation. Our experiments reveal an increase in recall values with increasing $\tau$, peaking at $\tau=0.01$. However, further increases in $\tau$ diminish performance, as non-informative patches are inadvertently included in the correspondence estimation. Ultimately, we select $\tau = 0.01$ as the default value for all experiments based on its optimal balance between increased recall and avoiding the inclusion of non-informative patches.

\noindent{\textbf{Different Backbones.}}
\textcolor{black}{DINOv2 \cite{oquab2023dinov2} stands out as a prominent Vision Foundational Model (VFM), proficient in tackling various vision challenges in its pre-initialized state. In Table \ref{tab:bb}, we explored different variants of DINOv2 as the backbone for extracting global features and patch tokens, while using our method for re-ranking. Comparative results were obtained with DINOv2 backbones which underscore the scalability of our re-ranking approach across other backbones. Notably, the performance of DINOv2 backbones was primarily influenced by attention scores that were not fine-tuned for place recognition tasks.}

\begin{table}[t]
\centering
\caption{\textcolor{black}{Ablation study on MSLS Validation with various backbone architectures.}}
\label{tab:bb}
\vspace{-2mm}
\resizebox{\linewidth}{!}{%
\begin{tblr}{
  width = \linewidth,
  colspec = {Q[465]Q[44]Q[123]Q[123]Q[146]},
  column{odd} = {fg=black},
  column{3} = {c},
  column{4} = {c,fg=black},
  column{5} = {c},
  cell{4}{2} = {c},
  cell{5}{2} = {c},
  hline{1-2,6} = {-}{},
}
Backbone Architecture &  & R@1  & R@5  & R@10 \\
DINOv2 ViT-S/14              &  & 80.2 & 83.4 & 85.9 \\
DINOv2 ViT-B/14              &  & 85.4 & 89.2 & 92.7 \\
DINOv2 ViT-L/14              &  & 87.7 & 90.6 & 93.0    \\
PlaceFormer (Ours)           &  & 89.9 & 94.3 & 95.4    
\end{tblr}
}
\vspace{-6mm}
\end{table}

\subsection{Visualization}
In Fig.~\ref{fig:viz_results}, we present a detailed case illustrating various retrieved images along with the attention maps and the key patches used for the retrieval process. Fig.~\ref{fig:viz_results} (a) shows a query image along with the top-3 retrieval using global retrieval (Fig.~\ref{fig:viz_results} (b)) and the top-3 results following re-ranking (Fig.~\ref{fig:viz_results} (c)). It can be seen that all of the top-3 retrieval using global descriptors are incorrect, while the re-ranking using multi-scale patches yields correct top-3 retrieval. Figs.~\ref{fig:viz_results} (d) and (f) provide insight into the attention mechanism, showcasing the attention map $A_L$ from the final layer of the transformer, along with the fused attention maps $A_{2L}$ and $A_{3L}$ for the query image and the closest matching reference image. The attention maps reveal that higher scores are assigned to task-relevant regions in the image like buildings, while dynamic objects like cars, cyclists, and the sky receive lower attention values. Figs~\ref{fig:viz_results} (e) and (g) illustrate the identification of key patches at various scales based on attention scores the query image and the closest matching reference image. The inliers estimated between these patches of different scales are depicted in Figs~\ref{fig:viz_results} (h), (i), and (j). The inliers illustrate that employing patches of multiple scales facilitates the identification of more correspondences compared to using patches of similar scales. The additional inliers, in turn, contribute to the improved re-ranking of images.

\vspace{-2mm}
\section{Conclusion}
\label{sec:conclusion}
This paper introduces PlaceFormer, a novel approach to place recognition employing the vision transformer. PlaceFormer leverages patch tokens extracted from the vision transformer, synthesizing patches of multiple scales through fusion. The amalgamation of these multi-scale patches yields superior image retrieval results compared to existing state-of-the-art methods, which typically operate with patches of similar sizes across diverse benchmark datasets. Notably, the incorporation of attention scores from the vision transformer enables the identification of task-relevant regions in the image. Consequently, only patches corresponding to these pertinent regions are retained, effectively reducing memory usage in PlaceFormer. The selective use of key patches further accelerates the correspondence estimation for re-ranking, contributing to the overall efficiency of the proposed approach. Nevertheless, our approach has certain limitations, particularly in matching latency when contrasted with methods employing neural networks for re-ranking. The computational intensity arises from estimating homography using RANSAC. As future research, we intend to explore the development of network models capable of efficiently matching patches of various scales in diverse combinations, hence enhancing the overall efficiency of the method.

\vspace{-1mm}
\bibliographystyle{IEEEtran}
\bibliography{references}
\end{document}